\documentclass[pdflatex,sn-basic,Numbered]{sn-jnl}

\usepackage{graphicx}
\usepackage{amsmath,amssymb,amsfonts}
\usepackage{booktabs}
\usepackage{xcolor}
\usepackage{tikz}
\usetikzlibrary{positioning,arrows.meta}
\usepackage{bookmark}

\newcommand{\vT}{v_{\mathrm{T}}}
\newcommand{\btheta}{\boldsymbol{\theta}}

\newcommand{\ms}{\,\mathrm{m/s}}

\newcommand{\GNPool}{51}
\newcommand{\GNTotal}{76}
\newcommand{\GNRej}{25}

\newcommand{\GMonoMed}{13.3}

\newcommand{\RNetPoolMedAbs}{8.6}

\newcommand{\MFPoolMedAbs}{9.9}

\begin{document}

\title[Monocular Badminton Speed Estimation]{Monocular Badminton Speed Estimation with Air Drag and Court Geometry}

\author[1]{\fnm{Diwen} \sur{Huang}}\email{diwenhuang42@gmail.com}
\equalcont{Lead author.\par$^{\ddagger}$Senior author.}
\author[2]{\fnm{Sarvesh} \sur{Madullapalli}}\email{sarveshm@ucla.edu}
\author[3]{\fnm{Ishan} \sur{Pendyala}}\email{ishanpen@stanford.edu}
\author[3]{\fnm{Oliver} \sur{Chen}}\email{ochen30@stanford.edu}
\author[4]{\fnm{Kevin} \sur{Li}}\email{kli19142@usc.edu}
\author[3]{\fnm{William} \sur{Wang}}\email{wwang28@stanford.edu}
\makeatletter\g@addto@macro\artauthors{$^{\ddagger}$}\makeatother

\affil[1]{\orgname{Independent Researcher}, \orgaddress{\city{Vancouver}, \state{BC}, \country{Canada}}}
\affil[2]{\orgname{University of California, Los Angeles}, \orgaddress{\city{Los Angeles}, \state{CA}, \country{USA}}}
\affil[3]{\orgname{Stanford University}, \orgaddress{\city{Stanford}, \state{CA}, \country{USA}}}
\affil[4]{\orgname{University of Southern California}, \orgaddress{\city{Los Angeles}, \state{CA}, \country{USA}}}

\abstract{A single camera cannot distinguish between a slowly moving object nearby and a rapidly moving object in the distance without additional information. For a shuttlecock, air drag provides a strong deceleration cue that can be combined with known court geometry. We fit a gravity-and-drag trajectory to smartphone video and compare a free monocular fit with a fit constrained to cross the net plane. Camera calibration uses court lines, and the trajectory is fitted to observed shuttle positions and frame timestamps by robust nonlinear least squares. We evaluated 51 badminton shots from two recording sessions, spanning drop shots, drives, half-smashes and smashes. The net-plane fit reduced the median absolute speed difference from a drag-based stereo reference from 13.3\% to 8.6\%. A second reference, obtained by differentiating triangulated positions without a drag model, also gave a smaller median difference for the net-plane fit. The results support combining court geometry with flight dynamics for badminton speed estimation from a single camera.}


\maketitle

\section{Introduction}
\label{sec:intro}

A single camera cannot distinguish between a slowly moving object nearby and a rapidly moving object in the distance~\citep{hartley2004multiple}. Known physical principles can provide scale information that a monocular camera cannot obtain from an isolated image. When the acceleration due to gravity is known, the visible curvature acts as a ruler~\citep{gravityheight}. This principle has been used to measure the trajectory of a soccer ball \citep{kim1998physics} and the general motion of projectiles \citep{ribnick2009estimating,pmgs}.

A shuttlecock slows rapidly after impact, making air drag an important part of its trajectory~\citep{cooke1999,phomsoupha2015}. The fractional loss rate is proportional to speed~\citep{cohen2015physics}. A calibrated camera records both the image trajectory and the elapsed time, allowing the trajectory to be fitted under a gravity-and-drag model. Court geometry supplies additional spatial information~\citep{farin2004robust,bwflaws}. Our aim is to assess how these sources of information can be used to estimate badminton speed from one camera.

\begin{figure}[t]
  \centering
  \includegraphics[width=0.65\linewidth]{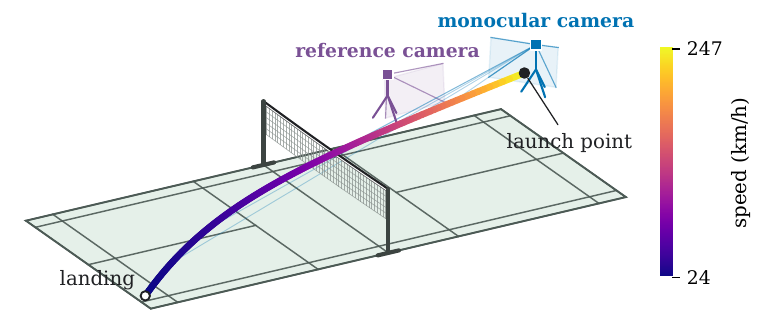}
  \caption{Filming setup and a reconstructed session-2 trajectory, colored by instantaneous speed. One view supplies the monocular track; the other cameras support reference measurements}
  \label{fig:schematic}
\end{figure}

We compare a free monocular fit with a court-constrained fit that describes the trajectory at its net crossing. A third condition fixes launch depth using stereo information to assess the effect of launch-position uncertainty. The experiments cover 51 shots across two players, two courts and two camera pairs. The central question is how much the court constraint improves speed estimation, and whether that improvement persists when the reference is constructed without a drag model.

MonoTrack fits quadratic drag to monocular badminton video using prior information on endpoint poses \citep{monotrack}. TT3D fits drag, Magnus lift and rebound to monocular table-tennis video via reprojection error \citep{tt3d}. A prior monocular pipeline was separately tested against a single fixed reference velocity \citep{huang2025system}. Here, we evaluate speed over a range of badminton shots and separate agreement with a physics-based stereo fit from agreement with a reference built from triangulated positions.

\section{Methods}
\label{sec:methods}

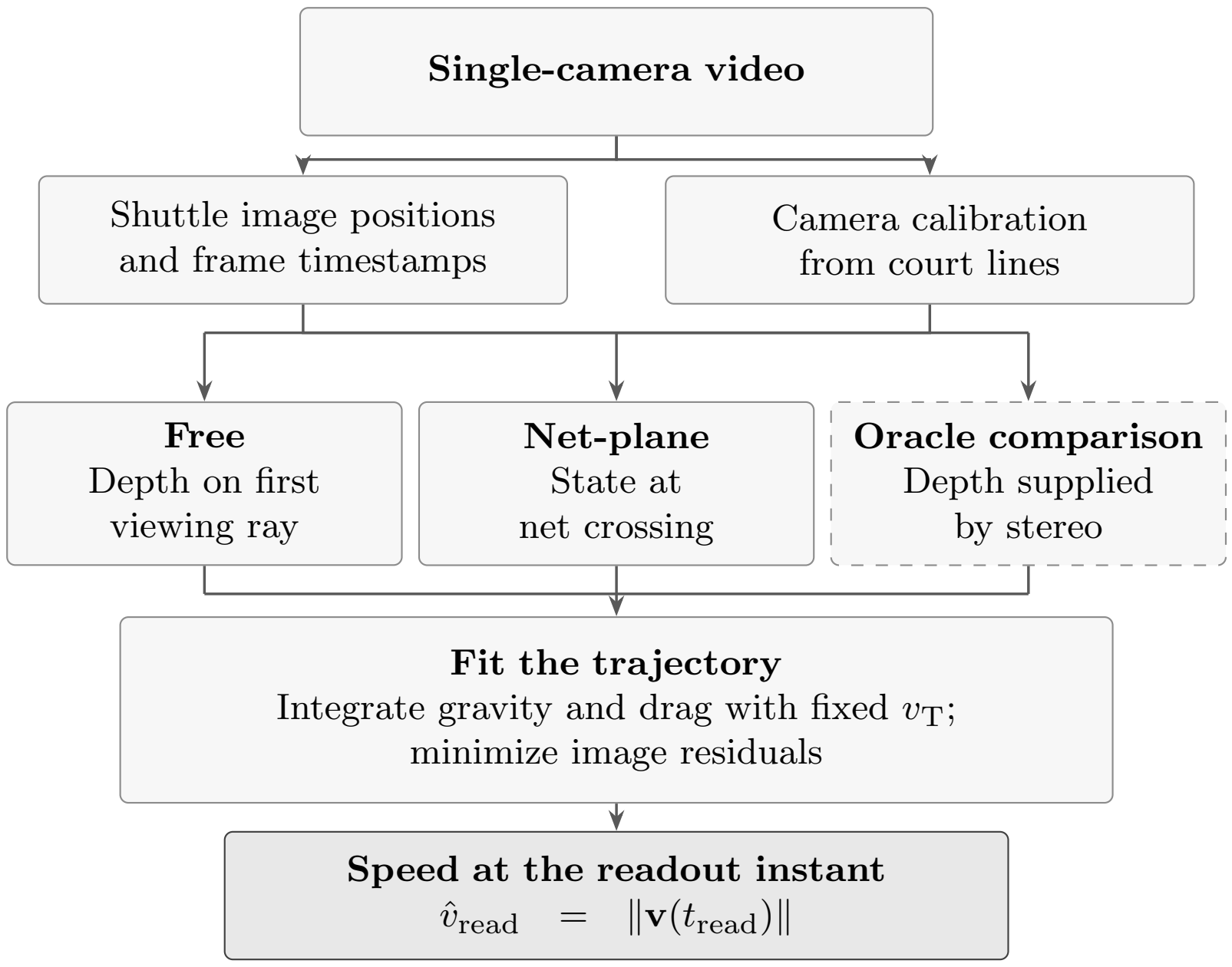
\begin{figure}[t]
\centering
\begin{tikzpicture}[
  font=\small, >={Stealth[length=1.8mm]},
  box/.style={draw=black!45, fill=black!3, rounded corners=2pt,
    align=center, inner sep=5pt, minimum height=11mm,
    execute at begin node={\hyphenpenalty=10000}},
  arr/.style={->, draw=black!65, line width=0.65pt},
  wire/.style={draw=black!65, line width=0.65pt}]
\node[box, text width=5.1cm] (video) at (0,0) {\textbf{Single-camera video}};
\node[box, text width=4.2cm] (track) at (-2.7,-1.45)
  {Shuttle image positions\\and frame timestamps};
\node[box, text width=4.2cm] (calib) at (2.7,-1.45)
  {Camera calibration\\from court lines};
\draw[arr] (video.south) -- ++(0,-0.2) -| (track.north);
\draw[arr] (video.south) -- ++(0,-0.2) -| (calib.north);
\node[box, text width=3.05cm] (free) at (-3.55,-3.55)
  {\textbf{Free}\\Depth on first viewing ray};
\node[box, text width=3.05cm] (net) at (0,-3.55)
  {\textbf{Net-plane}\\State at\\net crossing};
\node[box, dashed, text width=3.05cm] (oracle) at (3.55,-3.55)
  {\textbf{Oracle comparison}\\Depth supplied by stereo};
\draw[wire] (track.south) -- (-2.7,-2.25) -- (2.7,-2.25) -- (calib.south);
\draw[wire] (-3.55,-2.25) -- (3.55,-2.25);
\foreach \c in {free,net,oracle} {
  \draw[arr] (\c.north |- 0,-2.25) -- (\c.north);
}
\node[box, text width=8.2cm, minimum height=16mm] (fit) at (0,-5.5)
  {\textbf{Fit the trajectory}\\Integrate gravity and drag with fixed $v_{\mathrm{T}}$;\\minimize image residuals};
\foreach \c in {free,net,oracle} {
  \draw[wire] (\c.south) -- (\c.south |- 0,-4.5);
}
\draw[wire] (-3.55,-4.5) -- (3.55,-4.5);
\draw[arr] (0,-4.5) -- (fit.north);
\node[box, draw=black!75, fill=black!9, text width=6.4cm] (speed) at (0,-7.1)
  {\textbf{Speed at the readout instant}\\$\hat v_{\mathrm{read}}=\lVert\mathbf v(t_{\mathrm{read}})\rVert$};
\draw[arr] (fit.south) -- (speed.north);
\end{tikzpicture}
\caption{Trajectory estimation pipeline. Free and net-plane fits use a single camera; the oracle comparison additionally uses stereo launch depth}
\label{fig:pipeline}
\end{figure}

\subsection{Physical model}
\label{sec:background}
\label{sec:identifiability}

A shuttlecock loses speed rapidly after impact. We describe its flight using gravity and a drag force proportional to the square of its speed~\citep{cooke1999,chan2012}. This model follows the motion of the shuttle as a whole, represented by its position and velocity. The drag force is $-\tfrac12\rho_{\mathrm{air}}C_D A\lVert\mathbf{v}\rVert\mathbf{v}$, where $\rho_{\mathrm{air}}$ is air density, $C_D$ is the drag coefficient and $A$ is the reference area. For a shuttle of mass $m$, the forces balance at terminal speed: $mg=\tfrac12\rho_{\mathrm{air}}C_D A\vT^2$, giving the trajectory model
\begin{equation}
  \dot{\mathbf{p}}=\mathbf{v},\qquad
  \dot{\mathbf{v}}=\mathbf{g}-k\lVert\mathbf{v}\rVert\mathbf{v},\qquad
  k=\frac{g}{\vT^2}.
  \label{eq:ode}
\end{equation}
Here $\mathbf{p}$ and $\mathbf{v}$ are position and velocity, $\mathbf{g}$ is gravitational acceleration and $g=\lVert\mathbf{g}\rVert$. Based on the literature \citep{cohen2015physics,chan2012}, a value of $\vT=6.70\ms$ is adopted and used for all drag fits. Using terminal speed groups the aerodynamic constants into a single measurable parameter~\citep{farina2002,dodin2007ballistic}. We therefore do not need separate measurements of shuttle mass, reference area and drag coefficient. The same value of $\vT$ is used throughout, so differences between fitting conditions come from the information supplied to the estimator.

The role of drag is easiest to see in the fast part of the flight. For speed $v=\lVert\mathbf{v}\rVert$ well above $\vT$, the drag acceleration is much larger than gravity. Neglecting gravity for this approximation gives
\begin{equation}
  \frac{dv}{dt}\simeq-kv^2,\qquad
  v(t)\simeq\frac{v_0}{1+kv_0t},\qquad
  \frac{v(t)-v(t+\Delta t)}{v(t)}\simeq kv(t)\Delta t.
  \label{eq:decay}
\end{equation}
Here $v_0$ is the speed at the start of the interval. The last expression applies over a short time step, with $kv(t)\Delta t\ll1$. It shows why drag is useful for reconstruction: faster motion produces a larger fractional speed loss in the same elapsed time. Two trajectories with different physical speeds cannot retain the same decay profile simply by changing their distance from the camera~\citep{nardoneaidala1981,dodin2005fusion}.

Gravity supplies scale information through a different feature of the trajectory: its downward curvature. Both effects are included when fitting Equation~\ref{eq:ode}; the simplified decay law is used to explain the physical cue. In our recordings, the gravity-only model already constrains scale locally, but its estimated speeds are systematically lower than the drag-based reference. Including drag allows the estimator to represent the rapid slowing that occurs in real shuttle flight. The results section compares the two models directly.

\subsection{Trajectory estimation}
\subsubsection{From image observations to a flight trajectory}
\label{sec:estimator}
The input is a sequence of shuttle centers and their timestamps from one camera (Fig.~\ref{fig:pipeline}). Court lines provide the known dimensions needed to calibrate the camera and express the reconstructed trajectory in court coordinates~\citep{farin2004robust}. Given a candidate initial position and velocity, we integrate the physical model to predict where the shuttle will be at each observation time. Projection then maps those 3D positions back into the image:

\begin{equation}
  \hat{\mathbf{x}}_i=\pi\!\left(\mathbf{p}(t_i;\btheta)\right),\qquad
  \pi(\mathbf{p})=\operatorname{dehom}\!\left(\mathbf{K}\mathbf{R}(\mathbf{p}-\mathbf{C})\right),
  \label{eq:projection}
\end{equation}
where $\mathbf{K}$ contains the camera intrinsics, $\mathbf{R}$ rotates world coordinates into the camera frame and $\mathbf{C}$ is the camera center~\citep{hartley2004multiple}. The operation $\operatorname{dehom}$ divides the first two coordinates by the third. A single image point defines a viewing ray, leaving the distance along that ray unknown. In the free monocular fit, we use the first observed shuttle center to define this ray and write the initial position as
\begin{equation}
  \mathbf{p}_0=\mathbf{C}+s\mathbf{d},\qquad
  \btheta=(s,\mathbf{v}_0)\in\mathbb{R}^4,
  \label{eq:ray}
\end{equation}
where $\mathbf{d}$ is the unit viewing ray and $s$ is the unknown depth. The fit therefore estimates four quantities: depth and the three components of initial velocity. A change in depth changes the physical distance represented by the observed image motion, while a change in velocity changes both travel distance and deceleration. The optimizer adjusts them together to match the full observed track.

We integrate Equation~\ref{eq:ode} with fourth-order Runge--Kutta (RK4)~\citep{hairer1993solving} and estimate the initial state by minimizing image residuals:
\begin{equation}
  \hat{\btheta}=\underset{\btheta\in\Theta}{\arg\min}
  \sum_i\sum_{j=1}^{2}\ell_a\!\left(\hat{x}_{ij}(\btheta)-x_{ij}\right),\qquad
  \ell_a(r)=a^2\!\left(\sqrt{1+(r/a)^2}-1\right).
  \label{eq:objective}
\end{equation}
Here $\mathbf{x}_i$ is the observed shuttle center, and the two residual components correspond to horizontal and vertical image coordinates. The soft-$L_1$ loss, with $a=4$ pixels, reduces the influence of large image residuals~\citep{triggs2000bundle,virtanen2020scipy}. The set $\Theta$ contains the geometric and velocity bounds. We optimize from multiple starting states and retain the solution with the lowest objective; the bounds and starting-point checks are given in Supplement~\ref{app:freefisher}.

Timing enters the integration directly. We use each frame's presentation timestamp because smartphone recordings can have unequal intervals between frames. Reported speed is $\hat{v}_{\mathrm{read}}=\lVert\mathbf{v}(t_{\mathrm{read}};\hat{\btheta})\rVert$. We specify the common readout instant for the experiments below.

\subsubsection{Tracking and flight selection}
\begin{figure}[t]
\centering
\includegraphics[width=\linewidth]{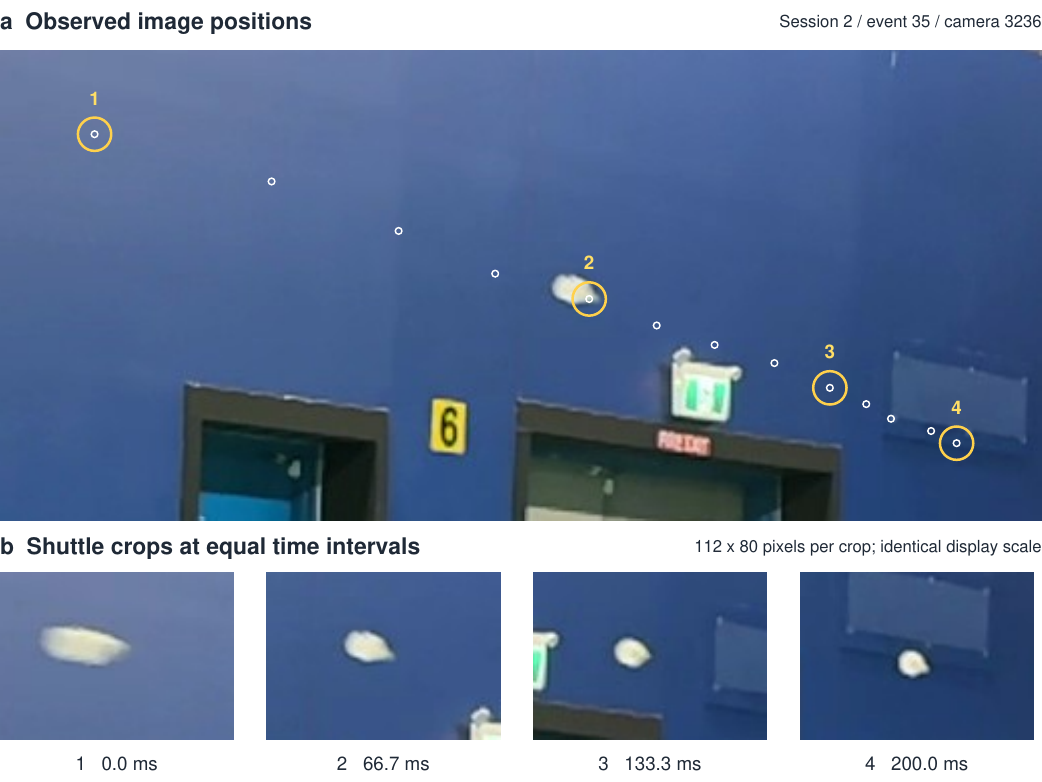}
\caption{Observed shuttle trajectory and corresponding image crops during a half-smash. Images are cropped to exclude players; image content was not modified.}
\label{fig:flightframes}
\end{figure}

A temporal detector trained on 22{,}000 manually labeled frames generates trajectories (F1 score of $0.86$, median error of $1.61$ pixels on held-out clips). The real-shot tracks undergo a manual outlier-correction pass. They are split at the frame of peak annotated image speed, used as an approximation to contact. Figure~\ref{fig:flightframes} illustrates the image observations using a cropped sequence from one evaluated event. The free-fall tracks described below were processed automatically. This gives a separate check of tracking and timing for motion with a known release height.

\subsubsection{Court constraint and oracle comparison}
\label{sec:anchor}
The free fit must infer depth from the observed motion. The badminton court offers an additional geometric constraint: a shot passing from one side to the other crosses the plane of the net. This crossing can be used to describe the trajectory without estimating the launch position from the player's body or racket.

We place the reference state at the net crossing and estimate its height, lateral position, velocity and time:
\begin{equation}
  \mathbf{p}(t_{\mathrm{net}})=(6.70,y,z)^{\mathsf T},\qquad
  \mathbf{v}(t_{\mathrm{net}})=(v_x,v_y,v_z)^{\mathsf T},\qquad
  \btheta_{\mathrm{net}}=(y,z,v_x,v_y,v_z,t_{\mathrm{net}}).
  \label{eq:net}
\end{equation}
Positions are in meters, with the net at court coordinate $x=6.70$\,m~\citep{bwflaws}. The drag ODE is integrated forward and backward from this state and fitted using the same image-residual objective (Equation~\ref{eq:objective}).

The crossing time is estimated together with the other state variables. Its search is restricted to the observed interval with a $50$\,ms margin at either end. All 51 evaluated events crossed the net plane inside the observation window. The fit uses six parameters, compared with four in the free fit, because its reference state includes the unknown crossing time. Thus the comparison evaluates a court-informed trajectory representation as a whole.

Both the free and net-plane conditions use only the monocular shuttle track and camera calibration. They differ in where the reference state is placed and how court geometry constrains it.

We use a third, oracle condition to examine the remaining uncertainty in launch depth. Here the stereo trajectory supplies depth on the first viewing ray, leaving velocity to be fitted from the monocular track. This provides a useful comparison for how much additional position information could improve the reconstruction. Unlike the first two conditions, the oracle requires stereo information.

Finally, we repeat each fitting condition with gravity alone. Keeping the geometry of each condition while removing drag lets us examine how the assumed dynamics affect the recovered speed.

\subsection{Experiments}
\subsubsection{Recordings}
\label{sec:stereo}
\label{sec:validation}
Both stereo rigs used two smartphones operating at 240\,fps with overlapping fields of view. Clock correction and calibration used clapping sounds and court-line intersections, respectively. Detected court intersections were manually verified against a court template. The shuttle position was annotated frame by frame in both views. A single drag trajectory was then fitted jointly to both cameras at their respective timestamps.

\begin{table}[t]
\centering
\caption{Recording sessions. The two sessions differed in player, court and camera pair. Both used a Victor Champion No.\,1 shuttle of manufacturer speed grade 78. Speed ranges are from the drag-based reference at the half-frame readout.}
\label{tab:sessions}
\begin{tabular}{lrrl}
\toprule
 & Recorded & Evaluated & Reference range (km/h)\\
\midrule
Session 1 & 10 & 10 & 69.7--323.7\\
Session 2 & 66 & 41 & 59.6--264.9\\
\bottomrule
\end{tabular}
\end{table}

The pooled statistics include only events where the joint stereo fit reproduces its manual labels in both views to below 25 pixels. Of the $\GNTotal$ recorded events, $\GNPool$ pass this gate ($\GNRej$ excluded). The threshold was set before monocular validation. The evaluated set contains 7 drives, 8 drop shots, 12 half-smashes and 24 smashes, with monocular observation windows of 408--917\,ms. Supplement~\ref{app:gate} gives the exclusions and acquisition details.

\subsubsection{Speed readout and reference measurements}
Shuttle speed changes throughout flight, so comparisons must evaluate speed at the same instant. We use the half-frame convention: speed is read half a frame before the first post-contact observation in the monocular camera. This approximates contact between frames. We also report speeds under the at-click convention, which places the initial state at the first observation. Supplement~\ref{app:framerate} explains how this timing choice interacts with the initial viewing ray.

We use two reference reconstructions. The first jointly fits the physical model to both camera views. Its extra viewpoint helps constrain depth, but it shares the drag model, fixed $\vT$ and one camera view with the monocular estimator. To examine the effect of those shared dynamics, we also construct a model-free reference: time-paired observations are triangulated into 3D points~\citep{hartley1997triangulation}, a polynomial is fitted to each coordinate, and its derivative gives velocity at the same readout time. The polynomial supplies a smooth position curve whose derivative can be evaluated at the readout instant. Drag, gravity and $\vT$ do not enter this reconstruction. Polynomial degree is selected from degrees 2--5 using leave-one-out position residuals~\citep{hastie2009elements}; the supplement gives the selection rules and their sensitivity.

A third camera in session 2 provides a separate check on shared observations. Paired with the wide reference camera, it forms a stereo pair that excludes the monocular view. The overlapping recordings provide 25 events for this comparison. Both camera pairs use the drag model, so this check examines the effect of the observations used to construct the reference.

\subsubsection{Comparison measure}
For each event, we compare the estimated speed with the reference speed at the common readout instant. We take the absolute difference, divide by the reference speed and report the median percentage across events. Table~\ref{tab:journalresults} presents both references side by side and gives 95\% bootstrap intervals from resampling events~\citep{efron1993introduction}. This keeps the comparison measure the same while making the choice of reference visible.

\subsubsection{Free-fall test}
\label{sec:drops}
We also test the physical model with a simpler motion whose distance can be measured directly: a shuttle released from rest. Tape-measured height and video timestamps allow predicted fall time to be compared with observed fall time, without a stereo trajectory fit. According to Equation~\ref{eq:ode}, a shuttle released from rest falls a distance~\citep{cohen2015physics}
\begin{equation}
  h(t)=\frac{\vT^2}{g}\ln\!\cosh\!\left(\frac{gt}{\vT}\right),\qquad
  t(h)=\frac{\vT}{g}\operatorname{arcosh}\!\left(\exp\!\left(\frac{gh}{\vT^2}\right)\right).
  \label{eq:fall}
\end{equation}
Here $h$ is downward displacement from the release point. The inverse expression predicts fall time directly from measured height.
We conducted 17 free falls, captured at 120\,fps, from heights of 2.16, 2.84 and 3.53\,m. Automatic tracking using presentation timestamps gave a mean absolute timing error of 2.1\% at the adopted terminal velocity, whereas the drag-free model had a $+16.6\%$ timing bias. The drop test provides a check of the adopted drag parameter and timing pipeline at low speeds.

\subsection{Ethics statement}
Both sessions were filmed with the informed consent of the players, who agreed to the recording and use of the derived annotations.

\section{Results}
\label{sec:failure}

\subsection{Monocular speed estimates}
The net-plane constraint improved agreement with the drag-based stereo reference: median absolute percentage difference fell from $\GMonoMed\%$ for the free fit to $\RNetPoolMedAbs\%$ for the net-plane fit. The improvement was present in both sessions (Table~\ref{tab:journalresults}). The oracle comparison gave a smaller difference, showing the benefit of additional launch-position information.

\begin{table}[t]
\centering
\caption{Median absolute speed difference from each reference (\%). All 51 evaluated events use the half-frame readout. Brackets are 95\% event-bootstrap intervals from 10,000 resamples. The model-free column uses the paired standard-error degree rule.}
\label{tab:journalresults}
\label{tab:gensummary}
\small
\begin{tabular}{lrr}
\toprule
 & Drag-based stereo & Model-free stereo\\
\midrule
Free & 13.3 [11.0, 14.9] & 21.2 [17.1, 24.8]\\
Net-plane & 8.6 [7.4, 9.9] & 18.4 [13.5, 26.0]\\
Oracle & 3.5 [2.1, 5.7] & 10.7 [7.1, 13.4]\\
\midrule
\multicolumn{3}{l}{Against drag-based stereo, by session}\\
Session 1: free / net / oracle & 11.7 / 6.4 / 1.6 & $n=10$\\
Session 2: free / net / oracle & 13.5 / 9.3 / 3.9 & $n=41$\\
\bottomrule
\end{tabular}
\end{table}

The model-free comparison follows the same ordering under the selected degree rule: net-plane fitting gives a smaller median difference than the free fit, and the oracle gives the smallest difference. The reduction for net-plane fitting is smaller in this comparison, with overlapping bootstrap intervals. The reference therefore affects the size of the observed benefit, even when the direction of the pooled comparison is unchanged.

The model-free and drag-based references themselves differ by a median of $\MFPoolMedAbs\%$. The third-camera check gives a median signed reference difference of $-2.50\%$ and preserves the free, net-plane, oracle ordering on its 25-event subset. The two checks complement each other by changing either the reconstruction model or the camera observations used for the reference. Full per-event results are given in Table~\ref{tab:genevents}.

\subsection{Effect of drag}
Omitting drag increases the median difference from the drag-based reference from 13.3\% to 36.7\% for the free fit and from 8.6\% to 35.2\% for the net-plane fit. The gravity-only free fit reads low on all 51 events. With oracle anchoring, the corresponding change is smaller, from 3.5\% to 7.4\%. The largest changes occur when depth must be inferred from the monocular track. When stereo information fixes launch depth, the choice of dynamics has a smaller effect on the fitted speed. This is consistent with the physical model: drag changes the link between observed deceleration, depth and velocity. These ablations use the drag-based reference.

\section{Discussion}
The results show how court geometry and a simple flight model can be used together for monocular badminton reconstruction. The net provides a fixed plane in court coordinates, while gravity and drag describe how the shuttle moves through that plane. Placing the reference state at the crossing gives the estimator a useful geometric condition without needing a launch point derived from player pose. The improvement against the drag-based reference appears in both recording sessions, despite their different players and camera arrangements.

The oracle comparison helps interpret the remaining difference. Supplying launch depth produces the closest agreement with both reference reconstructions. This suggests that better spatial information remains valuable even when the flight dynamics are modeled. In practical use, camera calibration, a clear view of the shuttle and a sufficiently long observed track all contribute to the information available for fitting. The truncation and anchor-sensitivity checks in the supplement examine these factors.

The principal validation issue is the reference speed. The drag-based stereo fit shares the estimator's dynamics, so agreement alone cannot establish absolute velocity accuracy. The model-free comparison addresses this dependency, but differentiating triangulated positions makes its speed estimate sensitive to polynomial degree, especially near the start of a track. Across the degree and readout variants, the oracle improves on the free fit in all 14 comparisons, and the full free, net-plane, oracle ordering holds in 10. These sensitivity results are reported in Supplement~\ref{app:modelfree}. An independent speed measurement across the tested range would provide the next direct test of accuracy.

The present experiments use manually corrected tracks from two players and one shuttle type. They establish the reconstruction comparison on those tracks; segmentation and performance with uncorrected detections remain to be evaluated separately. Contact timing also deserves attention because the shuttle decelerates quickly~\citep{texier2012shuttlecock}. In the half-frame fits, the initial state is assigned an earlier time while remaining on the first observed ray, coupling the timing and geometric assumptions. A time-consistent contact refit, followed by tests on additional shuttle types and recording conditions, would extend the present evaluation.

\section{Conclusions}
\label{sec:conclusion}
We presented a monocular badminton speed estimator based on gravity, quadratic drag and known court geometry. Representing the trajectory at the net crossing improved agreement with the drag-based stereo reference in both recording sessions and gave a smaller pooled median difference from the selected model-free reference than the free fit. The oracle comparison identified launch depth as a useful source of further improvement. Together, these results motivate using court geometry alongside physical dynamics when reconstructing fast shuttle flight from one camera.

\backmatter

\bmhead{Supplementary information}
The supplement cited in the text as Supplement S1--S6 and Tables S1--S3 follows the references.

\section*{Statements and Declarations}

\bmhead{Funding}
No funding was received for conducting this study.

\bmhead{Competing interests}
The authors have no relevant financial or non-financial interests to disclose.

\bmhead{Ethics approval and consent to participate}
Both sessions were filmed with the informed consent of the players, who agreed to the recording and use of the derived annotations.

\bmhead{Data availability}
Trajectories, calibration and annotations are available from the authors upon request. Raw video and audio are withheld to protect participants' privacy.

\bmhead{Code availability}
The code used for the analysis will be made publicly available.

\bmhead{Use of large language models}
Large language models assisted with data-processing scripts and manuscript editing. The research ideas are the authors' own. The authors are responsible for the methods, analysis, interpretation and final manuscript.

\bibliography{refs}

\clearpage
\appendix
\renewcommand{\thesection}{S\arabic{section}}
\renewcommand{\thetable}{S\arabic{table}}
\renewcommand{\thefigure}{S\arabic{figure}}
\renewcommand{\theHsection}{supp.\arabic{section}}
\renewcommand{\theHtable}{supp.\arabic{table}}
\renewcommand{\theHfigure}{supp.\arabic{figure}}
\setcounter{section}{0}
\setcounter{table}{0}
\setcounter{figure}{0}
\section*{Supplementary material}

\section{Acquisition and event selection}
\label{app:gate}
\label{app:budget}
Session 1 used a close-up iPhone 16 Pro at 720p and a wide-angle iPhone 17 Pro at 1080p. Calibration residuals were 1.2 and 4.7 pixels, respectively. The close-up view lost the shuttle 58--100\,ms after contact; the wide view supplied the monocular track. The session contained nine smashes and one drop shot. Seven smashes were returned before landing, shortening the observed flight.

Session 2 used cameras 6461 (wide-angle iPhone 17 Pro) and 3236 (iPhone 16 Pro), with 3236 as the monocular view. Camera 7008 recorded a subset simultaneously. Court-calibration residuals were 4--9 pixels. Both sessions used the same shuttle, a Victor Champion No.\,1 of speed grade 78. The event gate retains 10 of 10 session-1 events and 41 of 66 session-2 events. Of the 25 excluded events, 10 fail the residual threshold in the wide view, 7 in the monocular view and 8 in both. Since selection uses the drag-based reference's residual, the model-free comparison is also conditional on that selection.

Session-1 slow-motion recordings were converted to full-speed playback and retained variable timestamps. Session-2 timestamps were corrected to the container's presentation timestamps before the reported fits. The session-1 fitted clock offset was 189.5\,ms, compared with 177.9\,ms from the clap. Its event-to-event variation was below 0.7\,ms, but the source of the difference from the clap remains unresolved. In session 2, measured clap offsets differed by about 1\,ms across three simultaneous recordings.

Formal stereo-fit uncertainty was approximately 1--2\,km/h, with estimated calibration-scale uncertainty of 2--3\%. The earlier sensitivity budget gave reference uncertainties of approximately 3\% for full arcs and 9\% for short arcs. These are model-conditional estimates, not independently established bounds on true speed.

\section{Fitting constraints and timing}
\label{app:freefisher}
\label{app:framerate}
The free fit restricts launch depth to the portion of the first viewing ray inside a fixed court-coordinate box and at least one meter from the camera. The box allows roughly eight meters behind either baseline, nine meters sideways and six meters of height. Launch velocity is bounded by
$\mathbf{v}_0\in[-80,130]\times[-80,80]\times[-80,60]$\,m/s.
No evaluated half-frame fit activates these bounds. At the at-click convention, six fits reach a velocity bound; widening it changes the pooled median difference from 18.3\% to 18.8\%.

The net-plane fit searches crossing position over $y\in[-1.5,8.0]$\,m and $z\in[-0.5,6.0]$\,m, crossing velocity over $[-90,90]\times[-60,60]\times[-60,40]$\,m/s, and crossing time over $[-0.05,T+0.05]$\,s relative to the first observation. Eighteen starting points cover three times, three speeds and both court directions. These restrictions, and the six-parameter representation, differ from the free fit; the experiment does not isolate a single geometric constraint.

All 51 published fits were checked with 113--424 optimization starts per event and condition. For 49 events, the fits were stable to numerical tolerance. Two events had annotations changed after the frozen fit snapshot; using the updated annotations shifts the at-click free median from 18.3\% to 18.8\% and leaves the oracle median at 7.8\%. The manuscript retains the frozen results consistently.

\paragraph{Readout convention.}
The reported half-frame fits place the initial position on the first observed ray while assigning the state a time half a frame earlier. Thus changing the readout also changes the geometric constraint; it is not simply evaluating one unchanged trajectory at another time. The at-click convention aligns these two instants. Against the drag reference, the at-click median differences are 18.3\% for the free fit and 7.8\% for the oracle fit, compared with 13.3\% and 3.5\% under the half-frame convention.

For drag-dominated motion, a time offset $\delta$ gives
$v(\delta)=v_0/(1+kv_0\delta)$.
At the recorded speeds, a one-frame offset at 240\,fps changes speed by pooled medians of $-4.5\%$ or $+5.0\%$. Subsampling the 240\,fps annotations to lower rates also changes which observation defines the first ray. Those tests retain the original tracking and timestamp precision and do not validate native 30\,fps recordings.

\section{Model-free reference and degree sensitivity}
\label{app:modelfree}
Time-paired observations from the two reference views are triangulated into 3D points. A polynomial is fitted independently to each coordinate and differentiated at the common readout instant. Drag, gravity and terminal velocity do not enter. The method still shares calibration, observations and event selection with the drag-based reference.

Candidate degrees 2--5 require at least $d+3$ triangulated points. Each point is left out in turn and the three coordinates are fitted to the remaining points. The omitted-point 3D residual supplies the selection score. The initial rule selects the smallest degree within one marginal standard error of the minimum score~\citep{hastie2009elements}. The paired-SE variant uses the standard error of the paired score differences instead. The paired variant was introduced after inspecting the initial rule's output; it was not specified in advance. Table~\ref{tab:modelfree} retains both rules and fixed-degree alternatives.

\begin{table}[ht]
\centering
\small
\caption{Median absolute speed difference from the model-free reference (\%), all 51 events. Rows use the half-frame convention except where indicated. The paired-SE row is used in the main article.}
\label{tab:modelfree}
\begin{tabular}{lrrr}
\toprule
Degree rule & Free & Net-plane & Oracle\\
\midrule
Marginal-SE & 24.0 & 18.9 & 11.8\\
Paired-SE & 21.2 & 18.4 & 10.7\\
Paired-SE, at click & 27.3 & 17.0 & 14.9\\
Fixed degree 2 & 24.6 & 26.8 & 14.0\\
Fixed degree 3 & 16.8 & 17.0 & 6.6\\
Fixed degree 4 & 14.4 & 10.5 & 5.8\\
Fixed degree 5 & 16.8 & 10.9 & 12.0\\
\bottomrule
\end{tabular}
\end{table}

The initial rule selects degree 2 on 33 events. A runs test~\citep{wald1940test} rejects the degree-2 residual-sign sequence on 20 events, indicating temporal structure that the polynomial has not captured. Increasing degree improves the representation of changing acceleration but can make the endpoint derivative unstable. Across degrees 2--5, the median event-level speed spread is 29.0\%; it exceeds 20\% on 39 of 51 events. The pooled difference between the polynomial and drag references ranges from 8.2\% to 15.0\% across the tested choices. This sensitivity prevents interpreting their agreement as a bound on shared model bias.

Across seven degree variants and two readout conventions, the oracle result is better than the free result in all 14 comparisons. The full free, net-plane, oracle ordering holds in 10. The intervals in the main table summarize event resampling, not uncertainty from degree selection, calibration or shared observations.

\section{Third-camera comparison}
\label{app:heldout}
Camera 7008 was not used in the primary fits. Pairing it with camera 6461 produces a reference that excludes monocular camera 3236. The overlap contains 25 events passing the same reference gate in both pairs. Recording stopped before the fastest smashes, so this subset does not check that speed band.

At a common at-click instant, the held-out pair's reference differs from the primary reference by a median of $-2.50\%$ (95\% bootstrap interval $[-3.13,-1.33]$). Both pairs still use the same drag model and share camera 6461. This comparison tests sensitivity to the monocular observations, not independence from the model.

\begin{table}[ht]
\centering
\small
\caption{Median absolute speed difference on the same 25-event overlap (\%), at-click readout. The oracle uses its original launch-depth anchor in both columns.}
\label{tab:heldout}
\begin{tabular}{lrr}
\toprule
 & Primary pair & Held-out pair\\
\midrule
Free & 17.8 & 20.1\\
Net-plane & 8.9 & 10.4\\
Oracle & 6.4 & 8.6\\
\bottomrule
\end{tabular}
\end{table}

\section{Physical and numerical checks}
\label{app:drops}
\label{app:sim}
\label{app:fisher}
\label{app:sports}
\label{app:anchorvariants}
\label{app:distortion}
\label{app:state}
The free-fall experiment includes six drops at 2.16\,m, six at 2.84\,m and five at 3.53\,m. At the adopted terminal velocity, the mean absolute timing error is 2.1\%, maximum error 9.0\% and bias $+0.5\%$. The pooled fitted terminal velocity is $6.97\pm1.13\ms$ (sample standard deviation; SEM $0.27\ms$). The per-height estimates decrease with height, so the pooled SEM does not capture all possible departures from constant drag.

In the terminal-velocity sensitivity sweep over 41 session-2 events, the net-plane median difference from the drag reference was 21.6\%, 9.3\% and 5.3\% at $\vT=6.0$, $6.7$ and $7.4\ms$, respectively. The free and oracle conditions preferred different values. The adopted $6.70\ms$ came from the literature, not this sweep. Reference agreement under these changes cannot determine the true terminal velocity or true speed.

Simulation covered 252 combinations of viewpoint, frame rate, noise and speed, with 500 trials per combination. In the 148 combinations passing a predicted relative-speed uncertainty threshold of 5\%, the ratio of RMSE to the Cram\'er--Rao bound~\citep{kay1993fundamentals} was 0.94--1.08. Coverage was 94.5\% in this group and 93.6\% in the remaining combinations. These results check the estimator under its assumed noise and dynamics; they do not validate the real-data uncertainty model.

The free-fit Fisher analysis gives a median gravity-only to drag-aware speed-variance ratio of 1.23. Over the observed windows, gravitational deflection is already sufficient to constrain scale locally. This supports interpreting drag's principal contribution here as correcting model bias rather than eliminating scale uncertainty.

The truncation experiment fits progressively longer portions of 50 tracks, with velocity bounds widened. Shorter windows give less stable estimates. The product $kv_0T$ describes the accumulated drag within a flight, but does not reliably rank error across events. We do not use it as a universal threshold for accepting trajectories or transferring the method to other sports.

Early player-pose anchors~\citep{winter2009} were worse than the free fit in session 1. Loosening the oracle launch-depth constraint to $\pm0.30$\,m raised the pooled median difference from 3.5\% to 6.2\%. Net-plane residual bias varies with crossing height and session; the lens-distortion correction tested in session 2 did not remove it. Rolling shutter was not corrected~\citep{meingast2005geometric}. These checks identify geometry and timing as priorities for improving the reconstruction.

\clearpage
\section{Per-event results}
Table~\ref{tab:genevents} retains the original evaluated-event values. Statistics in the main article use unrounded fits. Event identifiers follow the recording sessions; the dagger distinguishes session-1 events.
\begin{table}[!ht]
  \caption{Per-event speeds (km/h), half-frame convention. $T$ is the observed monocular window; $kv_0T$ is the accumulated-drag parameter, not the fraction of speed lost. The oracle relaxation allows $\pm0.30$\,m in launch depth. Session-1 events are marked $\dagger$.}
  \label{tab:genevents}
  \centering
  \scriptsize
  \setlength{\tabcolsep}{4pt}\renewcommand{\arraystretch}{0.92}
  \begin{tabular}{llrrrrrrrr}
    \toprule
    Event & Shot & $T$ (ms) & $v/\vT$ & $kv_0T$ & Reference & Free & Oracle & Relaxed & Net \\
    \midrule
    31 & drop shot & 900 & 2.5 & 3.3 & \textbf{60} & 60 & 56 & 58 & 58 \\
    23 & drop shot & 796 & 2.8 & 3.3 & \textbf{68} & 66 & 63 & 65 & 66 \\
    12$^\dagger$ & drop shot & 905 & 2.9 & 3.8 & \textbf{70} & 84 & 85 & 84 & 88 \\
    30 & drop shot & 917 & 3.4 & 4.6 & \textbf{83} & 106 & 86 & 89 & 101 \\
    1 & drive & 479 & 3.6 & 2.5 & \textbf{86} & 73 & 66 & 68 & 78 \\
    13 & drive & 530 & 3.8 & 2.9 & \textbf{92} & 84 & 86 & 84 & 87 \\
    5 & drive & 521 & 3.9 & 3.0 & \textbf{94} & 105 & 101 & 105 & 97 \\
    2 & drive & 600 & 4.0 & 3.5 & \textbf{96} & 89 & 82 & 85 & 98 \\
    9 & drive & 580 & 4.2 & 3.5 & \textbf{100} & 104 & 98 & 102 & 100 \\
    8 & drive & 508 & 4.4 & 3.3 & \textbf{107} & 121 & 108 & 112 & 118 \\
    7 & drive & 446 & 4.6 & 3.0 & \textbf{111} & 108 & 115 & 111 & 107 \\
    18 & drop shot & 613 & 4.6 & 4.2 & \textbf{112} & 128 & 111 & 116 & 122 \\
    17 & drop shot & 613 & 4.8 & 4.3 & \textbf{115} & 130 & 115 & 121 & 124 \\
    28 & drop shot & 721 & 5.1 & 5.4 & \textbf{123} & 152 & 97 & 102 & 143 \\
    14 & drop shot & 558 & 5.1 & 4.2 & \textbf{124} & 124 & 140 & 133 & 135 \\
    35 & half-smash & 521 & 6.6 & 5.0 & \textbf{158} & 191 & 159 & 167 & 167 \\
    91 & smash & 684 & 6.8 & 6.8 & \textbf{163} & 205 & 169 & 177 & 196 \\
    36 & half-smash & 546 & 7.0 & 5.6 & \textbf{170} & 215 & 176 & 185 & 199 \\
    37 & half-smash & 538 & 7.1 & 5.6 & \textbf{172} & 206 & 176 & 184 & 197 \\
    47 & half-smash & 555 & 7.3 & 5.9 & \textbf{176} & 173 & 159 & 165 & 193 \\
    64 & smash & 742 & 7.5 & 8.1 & \textbf{181} & 206 & 183 & 191 & 196 \\
    41 & half-smash & 438 & 7.6 & 4.9 & \textbf{183} & 154 & 169 & 162 & 165 \\
    39 & half-smash & 546 & 7.7 & 6.1 & \textbf{185} & 242 & 183 & 194 & 204 \\
    57 & half-smash & 559 & 7.9 & 6.4 & \textbf{189} & 201 & 182 & 192 & 218 \\
    74 & smash & 525 & 8.0 & 6.1 & \textbf{192} & 233 & 196 & 208 & 237 \\
    43 & half-smash & 438 & 8.0 & 5.1 & \textbf{192} & 215 & 175 & 182 & 206 \\
    49 & half-smash & 525 & 8.2 & 6.3 & \textbf{198} & 242 & 214 & 227 & 223 \\
    52 & half-smash & 463 & 8.4 & 5.7 & \textbf{202} & 238 & 201 & 213 & 211 \\
    67 & smash & 613 & 8.5 & 7.7 & \textbf{206} & 224 & 196 & 208 & 216 \\
    69 & smash & 621 & 8.7 & 8.0 & \textbf{211} & 253 & 207 & 216 & 233 \\
    87 & smash & 488 & 8.9 & 6.4 & \textbf{215} & 214 & 206 & 214 & 236 \\
    45 & half-smash & 421 & 9.0 & 5.5 & \textbf{217} & 242 & 201 & 210 & 234 \\
    85 & smash & 592 & 9.1 & 7.9 & \textbf{219} & 280 & 232 & 247 & 278 \\
    90 & smash & 667 & 9.1 & 8.9 & \textbf{220} & 246 & 225 & 237 & 250 \\
    78 & smash & 438 & 9.3 & 6.0 & \textbf{225} & 277 & 226 & 237 & 271 \\
    83 & smash & 512 & 9.4 & 7.0 & \textbf{226} & 248 & 228 & 240 & 245 \\
    71 & smash & 467 & 9.6 & 6.5 & \textbf{231} & 265 & 223 & 234 & 276 \\
    84 & smash & 434 & 9.6 & 6.1 & \textbf{232} & 263 & 229 & 241 & 251 \\
    75 & smash & 684 & 9.8 & 9.8 & \textbf{236} & 271 & 239 & 253 & 258 \\
    58 & half-smash & 625 & 9.9 & 9.1 & \textbf{239} & 228 & 206 & 216 & 250 \\
    73 & smash & 521 & 10.6 & 8.1 & \textbf{255} & 290 & 238 & 250 & 294 \\
    92 & smash & 471 & 11.0 & 7.6 & \textbf{265} & 274 & 255 & 269 & 284 \\
    7$^\dagger$ & smash & 450 & 12.1 & 8.0 & \textbf{293} & 278 & 291 & 278 & 328 \\
    5$^\dagger$ & smash & 567 & 12.3 & 10.2 & \textbf{298} & 275 & 294 & 277 & 287 \\
    3$^\dagger$ & smash & 408 & 12.6 & 7.5 & \textbf{305} & 371 & 321 & 337 & 325 \\
    2$^\dagger$ & smash & 438 & 12.7 & 8.1 & \textbf{306} & 303 & 301 & 303 & 324 \\
    1$^\dagger$ & smash & 509 & 12.8 & 9.5 & \textbf{309} & 264 & 285 & 270 & 294 \\
    4$^\dagger$ & smash & 417 & 12.8 & 7.8 & \textbf{309} & 271 & 306 & 288 & 318 \\
    6$^\dagger$ & smash & 567 & 12.9 & 10.7 & \textbf{311} & 303 & 312 & 303 & 310 \\
    11$^\dagger$ & smash & 442 & 13.4 & 8.7 & \textbf{324} & 270 & 293 & 277 & 296 \\
    10$^\dagger$ & smash & 446 & 13.4 & 8.8 & \textbf{324} & 288 & 318 & 299 & 297 \\
    \bottomrule
  \end{tabular}
\end{table}

\end{document}